\documentclass[runningheads]{llncs}
\usepackage{graphicx}
\usepackage{amsmath,amssymb} % define this before the line numbering.
\usepackage{color}
\usepackage[width=122mm,left=12mm,paperwidth=146mm,height=193mm,top=12mm,paperheight=217mm]{geometry}
\usepackage[sort,nocompress]{cite}
\begin{document}
% \linenumbers
\pagestyle{headings}
\mainmatter

\title{A Siamese Long Short-Term Memory Architecture for Human Re-Identification} % Replace with your title

\titlerunning{A Siamese LSTM Architecture for Human Re-Identification}

\authorrunning{Rahul Rama Varior, Bing Shuai, Jiwen Lu, Dong Xu, and Gang Wang}

\author{Rahul Rama Varior$^{\dagger}$, Bing Shuai$^{\dagger}$, Jiwen Lu$^{\S}$, Dong Xu$^{\ddagger}$, and Gang Wang$^{\dagger,}$\thanks{Corresponding author.}}

\institute{$\dagger$ School of Electrical and Electronic Engineering, Nanyang Technological University\\
${\S}$ Department of Automation, Tsinghua University \\
$\ddagger$ School of Electrical and Information Engineering, University of Sydney\\
 \email{\{rahul004,bshuai001,wanggang\}@ntu.edu.sg} \ \ \ \ \email{lujiwen@tsinghua.edu.cn} \ \ \ \ \email{dong.xu@sydney.edu.au}}

\maketitle

\begin{abstract}
Matching pedestrians across multiple camera views known as human re-identification (re-identification) is a challenging problem in visual surveillance. In the existing works concentrating on feature extraction, representations are formed locally and independent of other regions. We present a novel siamese Long Short-Term Memory (LSTM) architecture that can process image regions sequentially and enhance the discriminative capability of local feature representation by leveraging contextual information. The feedback connections and internal gating mechanism of the LSTM cells enable our model to memorize the spatial dependencies and selectively propagate relevant contextual information through the network. We demonstrate improved performance compared to the baseline algorithm with no LSTM units and promising results compared to state-of-the-art methods on Market-1501, CUHK03 and VIPeR datasets. Visualization of the internal mechanism of LSTM cells shows meaningful patterns can be learned by our method. 
%\dots
\end{abstract}

\section{Introduction}

Matching pedestrians across multiple camera views which is known as human re-identification has gained increasing research interest in the computer vision community. This problem is particularly important due to its application in visual surveillance. Given a probe (query) image, the human re-identification system aims at identifying a set of matching images from a gallery set, which are mostly captured by a different camera. Instead of manually searching through a set of images from different cameras, automated human re-identification systems can save enormous amount of manual labor. However, human re-identification is a challenging task due to cluttered backgrounds, ambiguity in visual appearance, variations in illumination, pose and so on.

Most human re-identification works concentrate on developing a feature representation \cite{lomo,yangcolor2014,salientcolorECCV14,zhang2014novel} or learning a distance metric \cite{lomo,ZhenliShiyu_CVPR2013,eccv14prid}. With the recent advance of deep learning technologies for various computer vision applications, researchers also developed new deep learning architectures \cite{ejazdeep2015,mcpbc,li2014deepreid,gatedscnn,sicir,domainguided,yi2014deep} based on Convolutional Neural Networks (CNNs) for the human re-identification task. Most of these handcrafted features as well as learned features have certain limitations. When computing histograms or performing convolution followed by max-pooling operation for example, the features are extracted locally and thus are independent of those features extracted from other regions \cite{brainsegm}.

In this paper, we explore whether the discriminative capability of local features can be enhanced by leveraging the contextual information, i.e. the features from other regions of the image.
\begin{figure}[!t]
\begin{center}
\includegraphics[width=0.9\linewidth]{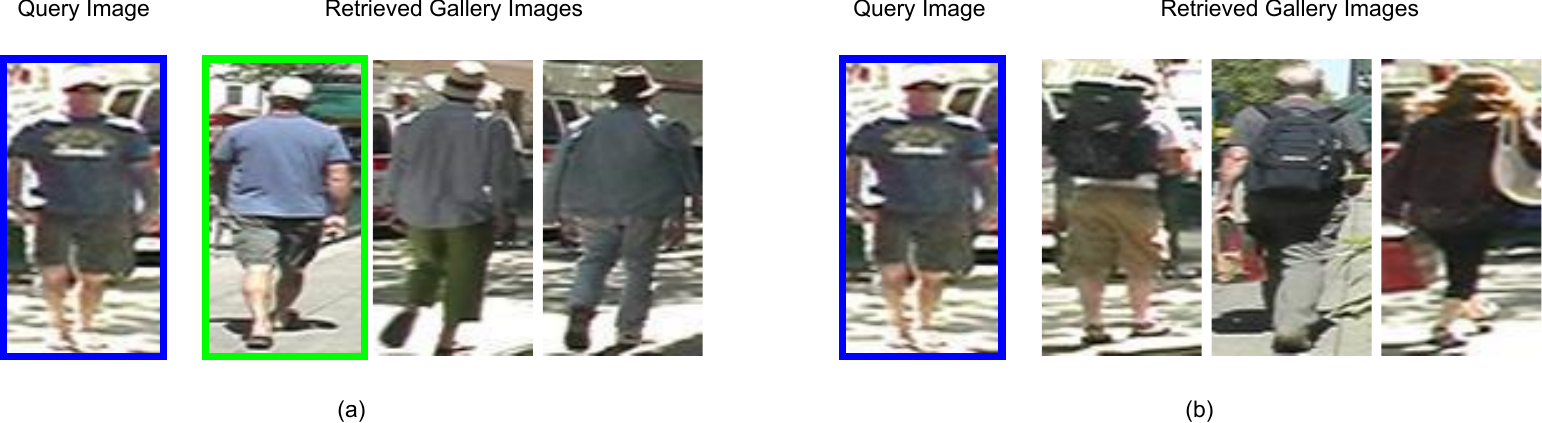}
\end{center}
   \caption{An example of the human re-identification scenario. (a) Result obtained by our framework (b) Result obtained by the baseline algorithm. Correct result retrieved as the best match is shown in green box. It can be observed that, more visually similar images were retrieved by the proposed approach (together with the correct match). Images are taken from the VIPeR dataset \cite{viper}. {\bf Best viewed in color.}}
\label{intro}
\end{figure}
Recurrent Neural Network (RNN) architectures have been developed to successfully model such spatial correlations \cite{zzcrnn,shuairnn} and adapt the local representations to extract more discriminative features. The self recurrent connections in RNNs enable them to learn representations based on inputs that it has previously `seen'. Thus the features learned by RNNs at any point can encode the spatial dependencies and `memorize' them. However, not all the spatial dependencies might be relevant for the image under consideration. Ideally, the network should be flexible to allow the propagation of certain contextual information that have discriminative capability and block the irrelevant ones. Thus the ambiguity of features can be reduced and more discriminative features can be learned. A variant of RNNs called Long Short-Term Memory (LSTM) \cite{lstm} cells have been used to spot salient keywords in sentences \cite{deepsentencelstm} and speech inputs \cite{keyword} to learn context (i.e., topic) relevant information. The advanced gating mechanisms inside the LSTM cell can regulate the information flowing into and out of the cell \cite{lstmsearch}. The extracted salient contextual information can further enhance the discriminative power of the learned local feature representations.

However, for human re-identification, in the embedded feature space, the feature vectors of similar pairs (i.e., from the same subject) must be `close' to each other while the feature vectors from dissimilar pairs should be distant to each other. 
To this end, we propose a siamese architecture based on LSTM cells. Siamese networks consist of two identical sub-networks joined at the output which are used for comparing two input fields \cite{siamese}. For learning the network parameters, inputs are therefore given in the form of pairs. The network is optimized by a contrastive loss function \cite{drlim}. The fundamental idea of the contrastive loss function is to `attract' similar inputs towards each other and `repel' dissimilar inputs. As a result, LSTM network can selectively propagate the contexts that can bring together the positive pairs and push apart the negative pairs.

The image is divided into several horizontal stripes and is represented as a sequence of image regions following \cite{zzcrnn} and starting from the first horizontal stripe, the LSTM cell progressively takes each of the horizontal stripes as inputs and decides whether to retain the information from the current input or discard it based on the information it captured from the current and previous inputs. Similarly, the LSTM cell can hide (or release) the contents of the memory from (or to) the other components of the network at each step. 
Detailed experimental evaluation of our proposed approach was conducted on three challenging publicly available datasets for human re-identification, Market-1501 \cite{market}, CUHK03 \cite{cuhk03} and VIPeR \cite{viper}. Our approach outperforms a baseline without LSTM units and achieve promising results compared to the state-of-the-art algorithms on all these datasets. We also provide intuitive visualizations and explanations to demonstrate the internal interactions of LSTM cells and prove the effectiveness of our approach. We summarize the major contributions of this paper as follows.

\begin{itemize}
\item We adapt the LSTM to human re-identification that can leverage the contextual information to enhance the discriminative capability of the local features. This significantly differs from the traditional methods that perform feature extraction locally and independent of other regions.
\item We propose a novel siamese LSTM architecture optimized by the contrastive loss function for learning an embedded feature space where similar pairs are closer to each other and dissimilar pairs are distant from each other.
\item Our approach achieves better performance when compared to a baseline algorithm (without LSTM units) as well as promising results when compared to several state-of-the-art algorithms for human re-identification. We also evaluate the multiplicative internal interactions of the LSTM cells and provide intuitive visualizations to demonstrate the effectiveness of our approach.
\end{itemize}
To the best of our knowledge, this is the first siamese architecture with LSTM as its fundamental component for human re-identification task.

%------------------------------------------------------------------------
\section{Related Works}
\label{related}
%-------------------------------------------------------------------------
\subsection{Human Re-Identification}

Most of the works on human re-identification concentrates on either developing a new feature representation \cite{lomo,yangcolor2014,salientcolorECCV14,zhang2014novel,bicovma2012,custompict,kviatkovsky2013color} or learning a new distance metric \cite{li2012human,pedagadilfda,ZhenliShiyu_CVPR2013,eccv14prid,lomo}. Color histograms \cite{eccv14prid,lomo,zhao2013unsupervised,zhao2013person}, Local Binary Patterns \cite{lbp,eccv14prid}, Color Names \cite{salientcolorECCV14,market}, Scale Invariant Feature Transforms \cite{lowe2004distinctive,zhao2013unsupervised,zhao2013person} etc are commonly used features for re-identification in order to address the changes in view-point, illumination and pose. In addition to color histograms, the work in \cite{lomo} uses a Scale Invariant Local Ternary Pattern (SILTP) \cite{siltp} features and computes the maximal local occurrence (LOMO) features along the same horizontal location to achieve view point invariance. Combined with the metric learning algorithm XQDA \cite{lomo}, LOMO features have demonstrated the state-of-the-art performance on both VIPeR \cite{viper} and CUHK03 \cite{cuhk03} datasets. But, all the above features are extracted locally and without considering the spatial context which can enhance the discriminative capability of the local representation. Different from the above works, our proposed approach concentrates on improving the discriminative capability of local features by modeling the spatial correlation between different regions within the image. However, we use the state-of-the-art features (LOMO \cite{lomo}) as the basic local features and further propose a new LSTM architecture to model the spatial dependency. Even though the proposed framework is optimized using the contrastive loss function \cite{drlim}, we would like to point out that any differentiable metric learning algorithms can be used to optimize the proposed siamese architecture.\\
\noindent{\bf Deep Learning for Human Re-Identification: } Research in deep learning has achieved a remarkable progress in recent years and several deep learning architectures have been proposed for human re-identification \cite{ejazdeep2015,mcpbc,li2014deepreid,gatedscnn,sicir,domainguided,yi2014deep}. The fundamental idea stems from Siamese CNN (SCNN) architectures \cite{siamese}. The first SCNN architecture proposed for re-identification \cite{yi2014deep} consists of a set of $3$ SCNNs for each part of the image. 
In \cite{li2014deepreid}, a convolutional layer with max-pooling is used to extract features followed by a patch matching layer which matches the filter responses across two views. A cross-input neighborhood difference module was introduced in \cite{ejazdeep2015} to learn the cross-view relationships of the features extracted by a 2-layer convolution and max-pooling layers. Cross-view relationships were also modeled in CNNs by incorporating matching gates \cite{gatedscnn} and cross-image representation subnetworks \cite{sicir}. Domain guided dropout was introduced for neuron selection in \cite{domainguided}. Multi-Channel part based CNN was introduced in \cite{mcpbc} to jointly learn both the global and local body-parts features. However, these architectures operate on convolved filter responses, which capture only a very small local context and is modeled completely independent of other regions. By using LSTM cells as the fundamental components in the proposed siamese architecture, we exploit the dependency between local regions for enhancing the discriminative capability of local features. Even though a recent work \cite{rcn} uses RNN for human re-identification, they use it to learn the interaction between multiple frames in a video and not for learning the spatial relationships.
%-------------------------------------------------------------------------

\begin{figure*}[!t]
\begin{center}
\includegraphics[width=0.8\linewidth]{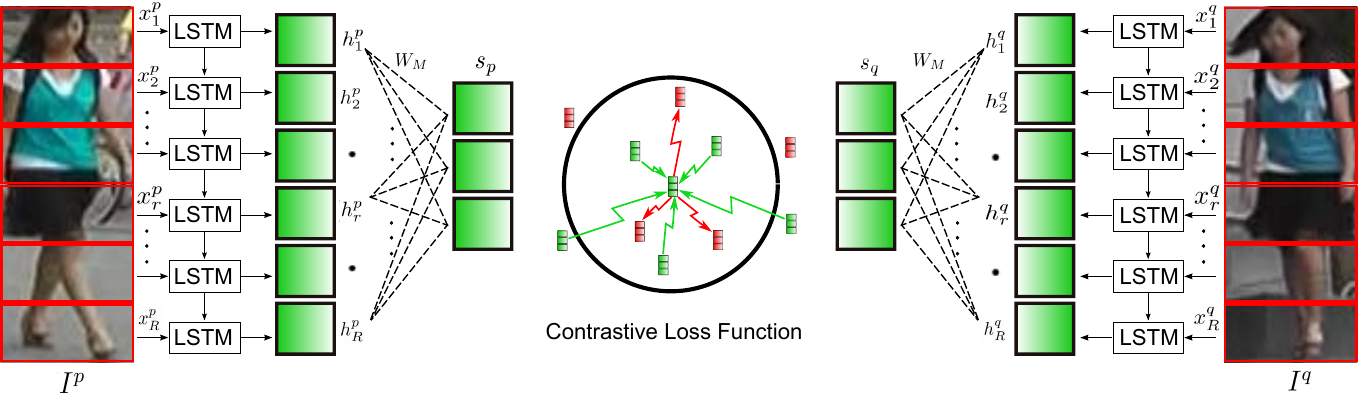}
\end{center}
   \caption{A diagram showing the proposed siamese LSTM architecture. The LSTM network initially processes the image features (${\bf x_r^{p}}$ and ${\bf x_r^{q}}$) sequentially to produce their hidden representations ${\bf h_r^{p}}$ and ${\bf h_r^{q}}$ respectively at each step. Once the hidden representations are obtained, a learned mapping (${\bf W_M}$) combines the hidden representations ${\bf h_r^{p}}$ and ${\bf h_r^{q}}$ to obtain ${\bf s_{p}}$ and ${\bf s_{q}}$ respectively. A contrastive loss function is used to compute the loss. Note that dividing the image into $6$  rows is merely for illustration and is not exactly the same in our experimental settings. {\bf Best viewed in color.}}
\label{framework}
\end{figure*}
\subsection{Recurrent Neural Networks}

Recurrent Neural Network (RNN) is a type of deep neural network that has recurrent connections, which enables the network to capture the context information in the sequence and retain the internal states. RNNs have achieved remarkable success in several natural language processing \cite{deepsentencelstm,seq2seq}, acoustic signal processing \cite{graves2014towards,acousticrnn}, machine translation \cite{seq2seq,karpathycaption} and computer vision \cite{shuairnn,zzcrnn,lstmscene,rcnnpinh} tasks. The fundamental idea behind RNN is that the connections with previous states enables the network to `memorize' information from past inputs and thereby capture the contextual dependency of the sequential data. Due to the difficulty in learning long range sequences (due to the vanishing gradient problem) \cite{Bengiolongterm}, Long Short Term Memory (LSTM) \cite{lstm} Networks were introduced and have been successfully applied to several tasks \cite{lstmscene,karpathycaption,deepsentencelstm}. In addition to capturing the contextual dependency, LSTM can also selectively allow or block the information flow through the network by using its advanced multiplicative interactions in the cell. Several researchers have conducted an empirical evaluation of different RNN architectures and provided intuitive explanations for the internal interactions in these architectures. For more details, we refer the interested readers to \cite{deepsentencelstm,karpathyweak,lstmsearch,empericallstm}. In \cite{deepsentencelstm,keyword}, it has been shown that LSTM cells can detect salient keywords relevant to a topic (context) from sentences or speech inputs. In \cite{brainsegm}, Pyramidal LSTM was proposed to segment brain images. However, the proposed work aims at building a siamese architecture with LSTM as its fundamental components for human re-identification. To the best of our knowledge, this is the first attempt to model LSTM cells in a siamese architecture for human re-identification.

\section{Our Framework}

\label{appraoch}

\noindent{\bf Overview:} The goal of our model is to match images of same pedestrians obtained from different surveillance cameras. The proposed siamese architecture consists of two copies of the Long-Short Term Memory network sharing the same set of parameters. The network is optimized based on the contrastive loss function. Figure \ref{framework} illustrates the proposed siamese LSTM architecture. Below, we explain our motivation through the introduction of the Long-Short Term Memory networks. Our proposed siamese architecture is further explained in detail with the optimization methodologies.
\subsection{Learning Contextual Dependency using LSTM}

RNN architectures have been previously used in \cite{zzcrnn,shuairnn,lstmscene} to model the spatial dependency and extract more discriminative features for image classification and scene labeling. While encoding such spatial correlations, traditional RNNs do not have the flexibility to selectively choose relevant contexts. Our work is motivated by the previous works \cite{deepsentencelstm,keyword} which have proven that the LSTM architectures can spot salient keywords from sentences and speech inputs. The internal gating mechanisms in the LSTM cells can regulate the propagation of certain relevant contexts, which enhance the discriminative capability of local features.
With these key insights, we propose a siamese LSTM network with pairs of images as inputs for the human re-identification task. The network is modeled in such a way that it accepts the inputs (horizontal stripes) one-by-one and progressively capture and aggregate the relevant contextual information. The contrastive loss function is used to optimize the network parameters so that the learned discriminative features can successfully bring the positive pairs together and push apart the negative pairs. Below, we explain the dynamics of a single layer LSTM architecture without peephole connections.
\subsubsection{Long Short-Term Memory Networks}
\label{rnn}

{LSTM }networks were introduced to address the vanishing gradient problem associated with RNNs and a more complete version was proposed in \cite{forgetgate,peephole,lstmbptt} with forget gate, peephole connection and full BPTT training. 
Mathematically, the update equations of LSTM cell at time $t$ can be expressed as
\begin{equation}
\label{lstmeq}
\begin{pmatrix}
{\bf i_t}\\ 
{\bf f_t}\\ 
{\bf o_t}\\ 
{\bf g_t}
\end{pmatrix} = \begin{pmatrix}
sigm\\ 
sigm\\ 
sigm\\ 
tanh
\end{pmatrix} {\bf W_{L}} \begin{pmatrix}
{\bf x_t}\\ 
{\bf h_{t-1}}
\end{pmatrix} 
\end{equation}
\begin{equation}
\label{celleq}
{\bf c_{t}} = {\bf f_t} \odot {\bf c_{t-1}} + {\bf i_t} \odot {\bf g_t} 
\end{equation}
\begin{equation}
\label{hiddeneq}
{\bf h_{t}} = {\bf o_t} \odot tanh({\bf c_{t}})
\end{equation}
From the above equations, it can be seen that the hidden representation ${\bf h_t} \in \mathbb{R}^{n}$ obtained at each step, Eq. (\ref{hiddeneq}), is a function of the input at the current time step (${\bf x_{t}} \in \mathbb{R}^{d}$) and the hidden state at the previous time step (${\bf h_{t-1}} \in \mathbb{R}^{n}$). We use ${\bf h_t}$ at each step as the feature representation in our framework. The bias term is omitted in Eq. (\ref{lstmeq}) for brevity. ${\bf W_{L}} \in \mathbb{R}^{4n \times (d+n)} $ denotes the LSTM weight matrix. Sigmoid ($sigm$) and hyperbolic tangent ($tanh$) are the non-linear activation functions which are applied element-wise. ${\bf c_t}\in \mathbb{R}^{n}$ denotes the memory state vector at time $t$. The vectors ${\bf i_t},{\bf o_t},{\bf f_t} \in \mathbb{R}^n$ are the $input$, the $output$ and the $forget$ gates respectively at time $t$ which modulates whether the memory cell is written to, reset or read from. Vector ${\bf g_t} \in \mathbb{R}^n$ at time $t$ is added to the memory cell content after being gated by ${\bf i_t}$. Thus the hidden state vector ${\bf h_{t}}$ becomes a function of all the inputs $\{ {\bf x_{1}}, {\bf x_{2}}, ..., {\bf x_{t}}\}$ until time $t$. The gradients of RNN are computed by back-propagation through time (BPTT) \cite{bptt}.\\
\noindent{\bf Internal Mechanisms: } The input gate can allow the input signal to {\it alter }the memory state or {\it block} it (Eq. (\ref{celleq})). The output gate can allow the memory contents to be {\it revealed} at the output or {\it prevent} its effect on other neurons (Eq. (\ref{hiddeneq})). Finally, the forget gate can update the memory cell's state by {\it erasing} or {\it retaining} the memory cell's previous state (Eq. (\ref{celleq})).  These powerful multiplicative interactions enable the LSTM network to capture richer contextual information as it goes along the sequence. 

\subsection{The Proposed Siamese LSTM Architecture}

For human re-identification, the objective is to retrieve a set of matching gallery images for a given query image. Therefore, we develop a siamese network to take pairs of images as inputs and learn an embedding, where representations of similar image pairs are closer to each other while dissimilar image pairs are distant from each other. All the images in the dataset are paired based on the prior knowledge of the relationships between the images (i.e., similar or dissimilar pairs). Consider an image pair ($I^p, I^q$) as shown in Figure \ref{framework}, corresponding to the $i^{th}$ pair of images ($i = \{1,2,..,N_{pairs}\}$). $N_{pairs}$ indicates the total number of image pairs. Let $Y^i \in [0, 1]$ be the label of the $i^{th}$ pair. $Y^i=0$ indicates that the images are similar and $Y^i=1$ indicates that they are dissimilar. Input features are first extracted from the images. Following previous works \cite{lomo,market,eccv14prid}, the input image is divided into several horizontal stripes and thus, treated as a spatial sequence as opposed to a temporal sequence typically observed in acoustic or natural language inputs. Dividing the image into horizontal rows has the advantage of translational invariance across different view points. We use $r$ as a suffix to denote the local features at a particular region (eg: ${\bf x_r}$; $r = \{1,2,...,R\}$). $R$ indicates the total number of regions (rows).

Dividing the image into rows has the advantage of translational invariance, which is important for human re-identification and is a commonly adopted strategy to represent image features \cite{lomo,market}. \\
\noindent{\bf Input Features: } We extract the state-of-the-art features (LOMO) and Color Names \cite{colornames} features from the images regions, corresponding to rows.
\begin{itemize}
\item{Local Maximal Occurrence (LOMO): } To extract the LOMO features, first, Color histogram  feature and SILTP features \cite{siltp} are extracted over $10 \times 10$ blocks with an overlap of 5 pixels. The feature representation of a row is obtained by maximizing the local occurence of each pattern at the same horizontal location. For an image with $128\times64$ pixels, this yields $24$ rows. Following the same settings, the features are extracted at $3$ different scales which resulted in 24, 11 and 5 rows per image.
\item{Color Names (CN) : } Features are extracted for $4\times4$ blocks with a step-size of $4$. Further a row-wise feature representation is obtained by combining the BoW features along the same horizontal location. We follow the same settings and codebook provided in \cite{market} for feature extraction. The final feature representation yields $16$ rows for each image with the size of $128\times64$.
\end{itemize}

Let the input features from the $r^{th}$ region (row) from the image pairs ($I^p, I^q$) be ${\bf x_r^p}$ and ${\bf x_r^q}$ respectively. As shown in Figure \ref{framework}, the input features ${\bf x_r^p}$ and ${\bf x_r^q}$ are fed into parallel LSTM networks. Each of these LSTM networks process the input sequentially and the hidden representations ${\bf h_r^p}$ and ${\bf h_r^q}$ at each step are obtained using Eq. (\ref{hiddeneq}). The hidden representation at a particular step is a function of the input at the current step and the hidden representation at the previous step. For example, ${\bf h_r^p}$ is a function of ${\bf x_r^p}$ and ${\bf h_{r-1}^p}$. In the proposed architecture, we use a single layer LSTM. Therefore, the hidden representations ${\bf h_r^p}$ and ${\bf h_r^q}$ obtained from the LSTM networks are used as the input representations for the rest of the network. 

Once the hidden representations from all the regions are obtained, they are combined to obtain ${\bf s_p}$ and ${\bf s_q}$ as shown below.
\begin{equation}
{\bf s_p} = {\bf W_M^T} [({\bf h_1^p})^T, ...,({\bf h_r^p})^T,...,({\bf h_R^p})^T]^T  ; \ r = {1,2,...,R}
\end{equation}
\begin{equation}
{\bf s_q} = {\bf W_M^T} [({\bf h_1^q})^T, ...,({\bf h_r^q})^T,...,({\bf h_R^q})^T]^T  ; \ r = {1,2,...,R}
\end{equation}
where ${\bf W_M} \in \mathbb{R}^{(R*n)\times (R*n)}$ is the transformation matrix.  $[.]^T$ indicates the transpose operator. The objective of the framework is that ${\bf s_p}$ and ${\bf s_q}$ should be closer to each other if they are similar and far from each other if they are dissimilar. The distance between the samples, ${\bf s_p}$ and ${\bf s_q}$ ($D_s({\bf s_p},{\bf s_q})$) can be given as follows:
\begin{equation}
D_s({\bf s_p},{\bf s_q}) = || {\bf s_p} - {\bf s_q}  ||_2
\end{equation}
Once the distance between the representations ${\bf s_p}$ and ${\bf s_q}$ is obtained, it is given as the inputs to the contrastive loss objective function. It can be formally written as:
\begin{multline}
\label{contrastivefinal}
%L({\bf s_p},{\bf s_q},Y^i; {\bf W_L}, W_M) = \\ (1-Y^i)\frac{1}{2}(D_s)^2 + (Y^i)\frac{1}{2} \{max(0, m - D_s )\}^2
L({\bf s_p},{\bf s_q},Y^i) = (1-Y^i)\frac{1}{2}(D_s)^2 + (Y^i)\frac{1}{2} \{max(0, m - D_s )\}^2
\end{multline}
where $m>0$ denotes a margin which acts as a boundary (with the radius $m$).  The intuition behind this loss function is that dissimilar pairs must be separated by a distance defined by $m$ and similar pairs must be as close as possible (i.e., distance tends to $0$). For more details regarding the loss function, we refer the interested readers to \cite{drlim}. The total loss can be obtained by taking the sum of the losses for all pairs . 

\noindent{\bf Network Training: } The overall loss is minimized so that similar pairs are closer to each other and dissimilar pairs are separated by $m$. The system is trained by back-propagating the gradients of Eq. (\ref{contrastivefinal}) with respect to ${\bf s_p}$ and ${\bf s_q}$ through the network. While generating the input image pairs, we do not consider all the negative images for a particular identity as it results in a biased dataset. Following the previous works \cite{ejazdeep2015}, the number of hard-negatives sampled is twice the number of positive pairs per image. To sample the hard-negatives, we consider the closest matching images in the input feature space (not in the raw image space). Even-though learning frameworks generalize better when trained with large datasets, we perform the training without any data augmentation or fine-tuning operations. \\
\noindent{\bf Optimization: }We use the mini-batch stochastic gradient descent method with the batch size of 100 pairs. Weight parameters (${\bf W_L}$ and ${\bf W_M}$) are initialized uniformly in the range of $[-a, a]$ where $a = \sqrt[]{1/(input\ size (d) + hidden\ size (n))}$ \cite{practicalnn}. In traditional RNN/LSTM architectures, gradients are computed using the BPTT algorithm \cite{bptt}. In the proposed siamese LSTM architecture, the gradients with respect to the feature vectors extracted from each pair of images are calculated and then back propagated for the respective branches independently using BPTT. As the parameters in each branch are shared, the gradients of the parameters are summed up and then the weights are updated. RMSProp \cite{rmsprop} per parameter adaptive update strategy is used to update the weight parameters. Following the previous works \cite{karpathycaption,karpathyweak}, we keep the decay parameter as $0.95$ and clip the gradients element-wise at $5$. These settings are fixed and found to be robust for all the datasets. The only tuned parameters were the hidden vector size ($n$), learning rate ($lr$) and margin ($m$) for the contrastive loss function (see Eq. (\ref{contrastivefinal})). Training is performed for a maximum of 20 epochs with an early stopping scheme if the cross-validation performance is found to be saturating. The optimal value for $m$ is tuned by cross-validation and is fixed to $0.5$ for all datasets. The optimal values for hidden size, learning rate as well as the learning rate decay coefficient (after every epoch) are dataset dependent. \\
\noindent{\bf Testing: } During the testing process, the local features for all the query and gallery images are extracted and mapped using the proposed framework to obtain the corresponding representations ${\bf s_p}$ ($p=\{1,...,N_{query}\}$) and ${\bf s_q}$ ($q=\{1,...,N_{gallery}\}$), where $N_{query}$ and $N_{gallery}$ denote the total number of images in the query set and the gallery set, respectively. The total number of query-gallery pairs will be $N_{query}\times N_{gallery}$. The final decision is made by comparing the Euclidean distance (i.e., matching scores) between all ${\bf s_p}$ and ${\bf s_q}$, $D_s({\bf s_p},{\bf s_q})$. When using multiple features, the matching scores obtained per query image with respect to all the gallery images for each feature are rescaled in the range of $0-1$ and then averaged. The final best match is the gallery image that has the least Euclidean distance based on the averaged scores.
\section{Experiments}
\label{expt}

In this section, we present a comprehensive evaluation of the proposed algorithm by comparing it with a baseline algorithm as well as several state-of-the-art algorithms for human re-identification. In most existing human re-identification works, the Cumulative Matching Characteristics (CMC) results were reported. However, in \cite{market}, human re-identification is treated mainly as a retrieval problem, so the rank 1 accuracy (R1 Acc) and the mean average precision (mAP) are used for performance evaluation. For a fair comparison with the baseline as well as the state-of-the-art algorithms, we report both CMC and mAP on all three datasets.\\
\noindent{\bf Baseline: } To evaluate the performance contribution of the proposed LSTM based siamese network, we implement a baseline method without using LSTM, i.e., with a mapping ${\bf W}$ alone. Features from all rows were concatenated and given as input in contrast to concatenating the hidden features from LSTM. Formally, the equations for obtaining ${\bf s_p}$ and ${\bf s_q}$ using a single layer baseline can be given as follows:
\begin{eqnarray}
\label{base}
{\bf s_p} = f({\bf W^T}[({\bf x_1^p})^T, ...,({\bf x_r^p})^T,...,({\bf x_R^p})^T]^T) \\ {\bf s_q} = f({\bf W^T} [({\bf x_1^q})^T, ...,({\bf x_r^q})^T,...,({\bf x_R^q})^T]^T)
\end{eqnarray}
where $f(.)$ is a non-linear activation function and ${\bf W}$ is the parameter matrix that is to be learned. The above system was optimized based on the same contrastive loss function in Eq. (\ref{contrastivefinal}). We also report the results using a multi-layer baseline which can be obtained by extending the above framework to multiple layers.

\subsection{Datasets and experimental settings}

The experiments were conducted on $3$ challenging human re-identification datasets, Market-1501 \cite{market}, CUHK03 \cite{cuhk03} and VIPeR \cite{viper}. 

\noindent {\bf Market-1501: } The Market-1501 dataset is currently the largest publicly available dataset for human re-identification with $32668$ annotated bounding boxes of $1501$ subjects. 
The dataset is split into $751$ identities for training and $750$ identities for testing as done in \cite{market}. We provide the multi-query evaluation results for this dataset. For multi-query evaluation, the matching scores for each of the query images from one subject are averaged.

\noindent {\bf CUHK03: } The CUHK03 dataset is a challenging dataset collected in the CUHK campus with $13164$ images of $1360$ identities from two camera views. Evaluation is usually conducted in two settings `labelled' with human annotated bounding boxes and `detected' with automatically generated bounding boxes. All the experiments presented in this paper use the `detected' bounding boxes as this is closer to the real-world scenario. 
Twenty random splits are provided in \cite{cuhk03} and the average results over all splits are reported. There are 100 identities for testing while the rest of the identities are used for training and validation. For multi-query evaluation, the matching scores from each of the query images belonging to the same identity are averaged. 

\noindent {\bf VIPeR: } The VIPeR dataset is another challenging dataset for human re-identification which consist of 632 identities captured from two cameras. For each individual, there is only one image per camera view. A stark change in illumination, pose and environment makes this dataset challenging for evaluating human re-identification algorithms. The dataset is split randomly into equal halves and cross camera search is performed to evaluate the algorithms. 

Table \ref{baseline} shows the performance comparison of the proposed algorithm with the baseline algorithm. It can be seen that the proposed LSTM architecture outperforms the single-layer and multi-layer baseline algorithms for all the datasets. Results indicate that feature selection based on the contextual dependency is effective for re-identification tasks. The Rank 1 performance on VIPeR dataset for the 3-layer baseline is lower compared to the $2$ layer and $1$ layer approach. We believe that this may be due to over-fitting as the dataset is smaller compared to the CUHK03 and Market-1501 datasets. Comparison with the state-of-the-art algorithms is shown in Table \ref{market}, \ref{cuhk03} and \ref{viper}. For a fair evaluation, we compare our results to only individual algorithms and {\bf not} to ensemble methods \cite{ensemble2015,zhao2014learning}. Some qualitative results are shown in Figure \ref{analysisfig}. It can be seen that the proposed algorithm retrieves visually similar images thereby improving the re-identification rate and mean average precision.

\subsection{Parameter tuning}
All the parameters are tuned by conducting cross-validation on the training data. In the supplementary material, we have shown the cross validation results using the LOMO \cite{lomo} features on the Market-1501 dataset. It was observed that when setting the LSTM hidden vector dimension larger than $25$, there is no significant improvement in the validation performance. Therefore, we set the hidden dimensions for Market-1501 dataset as $25$. Similarly, we observed that the optimal hidden dimensions for CUHK03 and VIPeR datasets were $50$. We also observed that the validation performance drops beyond margin $m = 0.75$. For our experiments, we set $m=0.5$ as there is a slight advantage in the validation performance. For tuning the learning rate, we conduct a log-uniform sampling in the range [$10^{-9}, \ 10^{-1}]$. For more detailed information on hyper-parameter search methods, we refer the interested readers to \cite{lstmsearch}.

\begin{table}[!t]

	\small
	\renewcommand{\arraystretch}{1}
	\setlength{\tabcolsep}{1.5pt}
	\caption{The CMC and mAP on the Market-1501, CUHK03 and VIPeR datasets.}
	\label{baseline}
	\centering
	\scalebox{0.7}{
	\begin{tabular}{|c|c|c|c|c|c|c|c|c|c|c|c|}
		\hline
		Dataset  & \multicolumn{2}{c|}{Market 1501} & \multicolumn{4}{c|}{CUHK03} & \multicolumn{4}{c|}{VIPeR} \\ 
		&  Rank 1 &  mAP   &  Rank 1 & Rank 5 &  Rank 10 & mAP   &  Rank 1 & Rank 5 &  Rank 10 & mAP  \\ 
		 \hline
Baseline (LOMO) - 1 Layer 	& {46.9}     	& 	21.3			& 49.1  & 76.0  & 85.3 &  40.1 					&  35.8  		& 62.3  & 75.0  & 42.8 \\
Baseline (LOMO) - 2 Layer 	& {47.8}     	& 	23.9			& 50.4 	& 77.6	& 85.9 &  40.9 					&  36.3  		& 63.6  & 75.3  & 42.9 \\
Baseline (LOMO) - 3 Layer 	& {48.4}     	& 	24.8			& 51.1	& 78.3	& 86.1 &  41.8 					&  34.8  		& 63.3  & 75.1  & 41.5 \\
{\bf LSTM (LOMO) - 1 Layer}  	& {\bf 51.8}     & {\bf 26.3}  	& {\bf 55.8} & {\bf 79.7}  & {\bf 88.2} & {\bf 44.2} & {\bf 40.5}   & {\bf 64.9}  & {\bf 76.3}  & {\bf 45.9}  \\
\hline 
Baseline (LOMO + CN) - 1 Layer  	&   52.1  & 27.1		&  51.6&  76.6  &  85.8 & 42.1 		&  36.1  & 64.9  & 75.6  & 43.0 \\
{\bf LSTM (LOMO + CN) - 1 Layer}  &{\bf 61.6}	  & {\bf 35.3}	& {\bf 57.3} & {\bf 80.1}   &{\bf 88.3}    & {\bf 46.3}		& {\bf42.4}   & {\bf  68.7}  & {\bf 79.4}  & {\bf 47.9}  \\
		 \hline
	\end{tabular}	}

\end{table}
\section{Analysis}
\label{analysis}

\subsection{Internal Mechanisms of the LSTM cell}

\begin{figure}[!t]
\begin{center}
\includegraphics[width=0.6\linewidth]{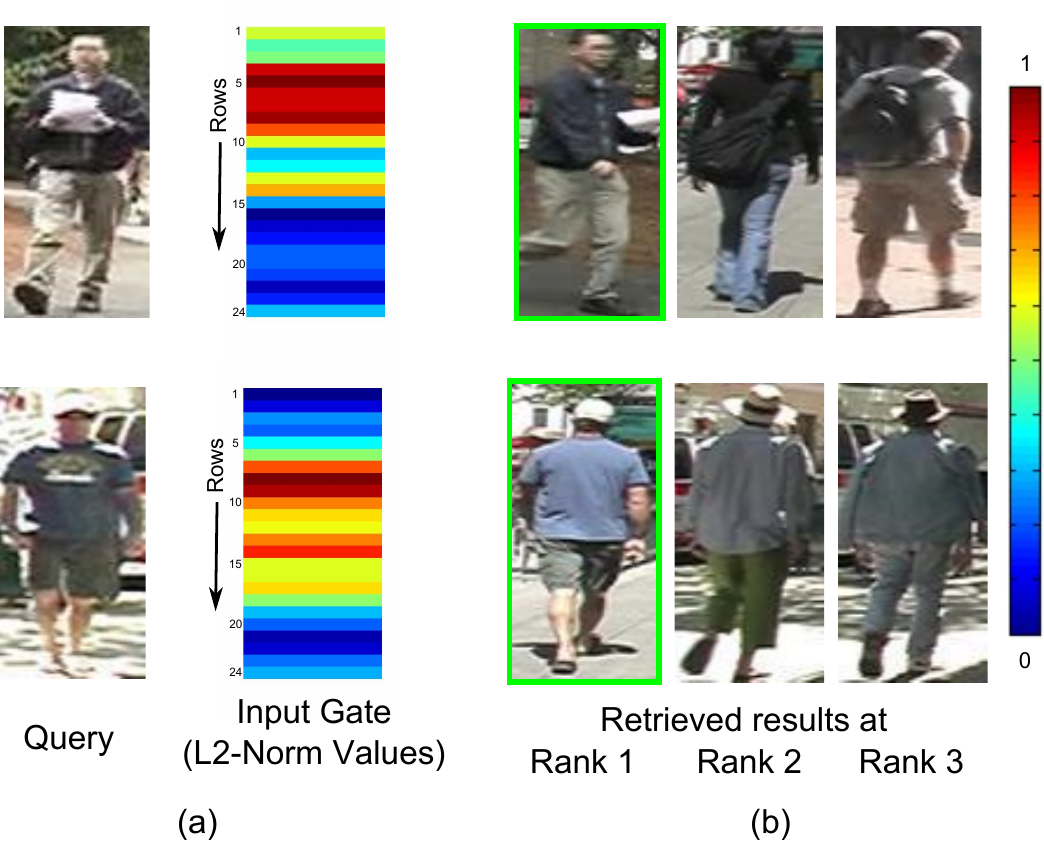}
\end{center}
   \caption{Qualitative Results: To clearly distinguish between the different values, the gate activations are given as {\bf heat maps.} (a) Two test queries and the $L2$ norm of the LSTM input gate activation values for the {\bf query} image. (b) Retrieved results for the query images. Images shown in green box are the correct match. See text for more details. {\bf Best viewed in color.}}
\label{analysisfig}
\end{figure}

Figure \ref{analysisfig} (a) shows two example queries from a testset of the VIPeR dataset and the input gate activations of the {\bf query} image. The retrieved matches for the query image are shown in Figure \ref{analysisfig} (b). From the `response' of the gates to certain inputs, we would like to answer the question whether the LSTM can select and `memorize' the relevant contextual information and discard the irrelevant ones. The optimal hidden dimensions ($n$) for the VIPeR dataset was found to be $50$ for the LOMO features. Therefore, the gate activations are $50$ dimensional vectors whose values range from $0-1$. In Figure \ref{analysisfig} (a), we show the L2 norm values of the gate activations at each step (24 steps in total corresponding to 24 local regions of the image). The L2 norm values are represented as a heat map where values close to $1$ (right saturated - information is propagated) is represented by darker shades of red and values closer to $0$ (left saturated - information is blocked) by deeper shades of blue. Kindly note that the $L2$ norm value is merely for better illustration and is not actually used in the system.

\noindent{\bf Input Gate: }The input gate values evolve in such a way that the relevant contextual information is propagated and unimportant ones are attenuated. For example, in the Figure \ref{analysisfig}(a), for the first query image, we can see that the representations from top $3$ rows, which mostly contains the background and head portion are attenuated with lower input gate activation values as the information is irrelevant to the context of the image (i.e., the visual appearance of the identity in this case). However, rows $4-9$ which mostly contains information from upper part of the body are selected and propagated.
For more step-by-step explanation of the internal mechanisms, we refer the readers to the supplementary material.\\

\subsection{State-of-the-art Comparison}
Table \ref{market}, \ref{cuhk03} and \ref{viper} shows the state-of-the-art comparisons on the Market-1501, CUHK03 and VIPeR datasets respectively. For Market-1501 dataset, a recent metric learning approach \cite{dns} outperforms ours. However, we believe that it can be complementary to our approach as the main contribution in this paper is on the feature learning aspect. For CUHK03 dataset, compared to other individual approaches, ours achieve the best results at Rank 1. For VIPeR dataset, several recent approaches \cite{sssvm,mcpbc,hgdreid,dns,scsp} outperform our approach. We believe that the reason is lack of positive pairs per image (only 1) and also the lack of total number of distinct training identities compared to other larger datasets. However, to improve the performance on feature learning approaches such as ours, transfer learning from larger datasets or data-augmentation can be employed. 
\begin{table}[!t]

	\scriptsize       
	\renewcommand{\arraystretch}{1}
	\setlength{\tabcolsep}{2pt}
	\caption{Performance Comparison of state-of-the-art algorithms for the Market-1501 dataset. Results for \cite{sdalf} and \cite{zhao2013unsupervised} are taken from \cite{market}.}
	\label{market}
	\centering
	{
	\begin{tabular}{|c|c|c|c|c|c|}
		\hline
		\bfseries Method & \bfseries Rank 1 & \bfseries mAP 
		\\\hline\hline			
		SDALF \cite{sdalf} & 20.53 & 8.20 \\
		eSDC \cite{zhao2013unsupervised} & 33.54 & 13.54 \\
		BoW + HS\cite{market} & 47.25 & 21.88 \\
		DNS \cite{dns} & {\bf 71.56} & {\bf 46.03} \\
		\hline		
		{\bf \begin{tabular}{@{}c@{}} Ours \end{tabular}} &{61.60}	  & {35.31}  \\
		\hline
	\end{tabular}	}
\end{table}
\begin{table}[!t]

	\scriptsize       
	\renewcommand{\arraystretch}{1.2}
	\setlength{\tabcolsep}{2pt}
	\caption{Performance Comparison of state-of-the-art algorithms for the CUHK03 dataset.}
	\label{cuhk03}
	\centering
	{
	\begin{tabular}{|c|c|c|c|c|}
		\hline
		\bfseries Method & \bfseries Rank 1 & \bfseries Rank 5 & \bfseries Rank 10 
		\\\hline\hline
		SDALF \cite{sdalf} & 4.9 & 21.0 & 31.7   \\

		ITML \cite{itml} & 5.14 & 17.7 & 28.3    \\

		LMNN \cite{Weinberger2009LMNN} & 6.25 & 18.7 & 29.0   \\

		eSDC \cite{zhao2013unsupervised} & 7.68 & 22.0 & 33.3    \\

		LDML \cite{isthatyou} & 10.9 & 32.3 & 46.7    \\

		KISSME \cite{kissmecvpr12} & 11.7 & 33.3 & 48.0     \\

		FPNN \cite{cuhk03} & 19.9 & 49.3 & 64.7   \\

		BoW \cite{market} & 23.0 & 45.0 & 55.7  \\
		BoW + HS \cite{market} & 24.3 & \_ & \_  \\
 		ConvNet \cite{ejazdeep2015} 	& 45.0	& 75.3 & 55.0   \\
		LX \cite{lomo} & 46.3 &  78.9 & {88.6}  \\
		MLAPG  \cite{mlapg} & 51.2 &  {83.6} & {92.1} \\
		SS-SVM \cite{sssvm} & 51.2 & 80.8 & 89.6 \\
		SI-CI \cite{sicir} & 52.2 & 84.3 & 92.3 \\
		DNS \cite{dns} & 54.7 & {\bf 84.8} & {\bf 94.8} \\
		
		\hline
%		{\bf \begin{tabular}{@{}c@{}} Ours \end{tabular}} & {40.51}   & { 73.0}  & {  84.6}  & { 45.18} \\
		{\bf \begin{tabular}{@{}c@{}} Ours \end{tabular}} & {\bf 57.3} & {80.1}   &{ 88.3}  \\
		\hline
	\end{tabular}	}
\end{table}
\begin{table}[!th]

	\scriptsize       
	\renewcommand{\arraystretch}{1.2}
	\setlength{\tabcolsep}{2.5pt}
	\caption{Performance Comparison of state-of-the-art algorithms using an individual method for the VIPeR dataset.}
	\label{viper}
	\centering
	{
	\begin{tabular}{|c|c|c|c|c|c|}
		\hline
		\bfseries Method & \bfseries Rank 1 & \bfseries Rank 5 & \bfseries Rank 10 
		\\\hline\hline
		LFDA \cite{pedagadi2013local} & 24.1 & 51.2 & 67.1   \\

		eSDC \cite{zhao2013unsupervised} & 26.9 & 47.5 & 62.3    \\

		{\begin{tabular}{@{}c@{}}Mid-level \cite{zhao2014learning}\end{tabular}}  & 29.1 & 52.3 & 65.9     \\

		SVMML \cite{ZhenliShiyu_CVPR2013} & 29.4 & 63.3 & 76.3    \\

		VWCM \cite{zhang2014novel} & 30.7 & 63.0 & 76.0    \\

		SalMatch \cite{zhao2013person} & 30.2 & 52.3 & 65.5    \\
		
		QAF \cite{queryadapt} & 30.2 & 51.6 & 62.4  \\
		
		SCNN \cite{yi2014deep} & 28.2 & 59.3 & 73.5   \\

 		ConvNet \cite{ejazdeep2015} 	& 34.8	& 63.7 & 75.8   \\

		CMWCE \cite{yangcolor2014} & 37.6 & 68.1 & {81.3}   \\

		{\begin{tabular}{@{}c@{}}SCNCD \cite{salientcolorECCV14} \end{tabular}} &  37.8 & 68.5 & 81.2   \\

		LX \cite{lomo} & 40.0 & 68.1 & 80.5   \\
		PRCSL \cite{prcsl} & 34.8 & 68.7 &	{82.3} \\
		MLAPG  \cite{mlapg} & 40.7	&	69.9 &	{82.3}\\
		MT-LORAE \cite{mtlorae} & 42.3 & {72.2} & 81.6 \\
		Semantic Representation \cite{transfsem} & 41.6 & 71.9  & {86.2}\\

		DGDropout \cite{domainguided} & 38.6 & \_ & \_ \\
		SI-CI \cite{sicir} & 35.8 & 67.4 & 83.5 \\

		SS-SVM \cite{sssvm} & 42.7 & \_	&	84.3 \\
		MCP-CNN \cite{mcpbc} & 47.8 & 74.7 & 84.8 \\
		HGD \cite{hgdreid} & 49.7 & 79.7	&	88.7 \\
		DNS \cite{dns} & 51.7 & 82.1 & 90.5 \\		
		SCSP \cite{scsp} & {\bf 53.5} & {\bf 82.6} & {\bf 91.5} \\
		
%		Gated S-CNN \cite{gatedscnn} & {37.8} & {66.9}   &{77.4} \\
		\hline		

		{\bf \begin{tabular}{@{}c@{}} Ours \end{tabular}} & {42.4}   & {68.7}  & {79.4}   \\
		\hline

	\end{tabular}	}
\end{table}
\section{Conclusion and Future Works}
\label{conclusion}

We have introduced a novel siamese LSTM architecture for human re-identification. Our network can selectively propagate relevant contextual information and thus enhance the discriminative capacity of the local features. To achieve the aforementioned task, our approach exploits the powerful multiplicative interactions within the LSTM cells by learning the spatial dependency. By examining the activation statistics of the input, forget and output gating mechanisms in the LSTM cell, we show that the network can selectively allow and block the context propagation and enable the network to `memorize' important information. Our approach is evaluated on several challenging real-world human re-identification datasets and it consistently outperforms the baseline and achieves promising results compared to the state-of-the-art.
\subsubsection*{Acknowledgments: }The research is supported by Singapore Ministry of Education (MOE) Tier 2 ARC28/14, and Singapore A*STAR Science and Engineering Research Council PSF1321202099. 
This research was carried out at the Rapid-Rich Object Search (ROSE) Lab at Nanyang Technological University. The ROSE Lab is supported by the National Research Foundation, Singapore, under its Interactive Digital Media (IDM) Strategic Research Programme. 
We thank NVIDIA Corporation for their generous GPU donation to carry out this research.

\bibliographystyle{splncs03}
\bibliography{egbib}
\end{document}